\documentclass[conference]{IEEEtran}
\IEEEoverridecommandlockouts
\usepackage{times}
\usepackage{latexsym}
\usepackage{hyperref}

\usepackage{cite}
\usepackage{amsmath,amssymb,amsfonts}
\usepackage{algorithmic}
\usepackage{graphicx}
\usepackage{textcomp}
\usepackage{xcolor}
\usepackage{latexsym}
\usepackage{blindtext}
\usepackage{multirow}
\usepackage[multiple]{footmisc}
\usepackage{mathtools}

\def\BibTeX{{\rm B\kern-.05em{\sc i\kern-.025em b}\kern-.08em
    T\kern-.1667em\lower.7ex\hbox{E}\kern-.125emX}}

\begin{document}

\title{\textsc{GAE-ISumm}: Unsupervised Graph-Based Summarization of Indian Languages}
% {\footnotesize \textsuperscript{*}Note: Sub-titles are not captured in Xplore and
% should not be used}
% \thanks{Identify applicable funding agency here. If none, delete this.}
% }

\author{
\IEEEauthorblockN{Lakshmi Sireesha Vakada\textsuperscript{1}, Anudeep Ch\textsuperscript{1},  Mounika Marreddy\textsuperscript{1}, Subba Reddy Oota\textsuperscript{2},
Radhika Mamidi\textsuperscript{1}}
\IEEEauthorblockA{\textsuperscript{1} IIIT Hyderabad, \textsuperscript{2} Inria Bordeaux, France \\
\small\{lakshmi.sireesha, mounika.marreddy\}@research.iiit.ac.in, anudeepch528@gmail.com, subba-reddy.oota@inria.fr
  and radhika.mamidi@iiit.ac.in}
}

% \author{\IEEEauthorblockN{1\textsuperscript{st} Given Name Surname}
% \IEEEauthorblockA{\textit{dept. name of organization (of Aff.)} \\
% \textit{name of organization (of Aff.)}\\
% City, Country \\
% email address or ORCID}
% \and
% \IEEEauthorblockN{2\textsuperscript{nd} Given Name Surname}
% \IEEEauthorblockA{\textit{dept. name of organization (of Aff.)} \\
% \textit{name of organization (of Aff.)}\\
% City, Country \\
% email address or ORCID}
% \and
% \IEEEauthorblockN{3\textsuperscript{rd} Given Name Surname}
% \IEEEauthorblockA{\textit{dept. name of organization (of Aff.)} \\
% \textit{name of organization (of Aff.)}\\
% City, Country \\
% email address or ORCID}
% \and
% \IEEEauthorblockN{4\textsuperscript{th} Given Name Surname}
% \IEEEauthorblockA{\textit{dept. name of organization (of Aff.)} \\
% \textit{name of organization (of Aff.)}\\
% City, Country \\
% email address or ORCID}
% \and
% \IEEEauthorblockN{5\textsuperscript{th} Given Name Surname}
% \IEEEauthorblockA{\textit{dept. name of organization (of Aff.)} \\
% \textit{name of organization (of Aff.)}\\
% City, Country \\
% email address or ORCID}
% \and
% \IEEEauthorblockN{6\textsuperscript{th} Given Name Surname}
% \IEEEauthorblockA{\textit{dept. name of organization (of Aff.)} \\
% \textit{name of organization (of Aff.)}\\
% City, Country \\
% email address or ORCID}
% }

\maketitle

\begin{abstract}
%\textcolor{red}{should also mention Indian languages}
Document summarization aims to create a precise and coherent summary of a text document. Many deep learning summarization models are developed mainly for English, often requiring a large training corpus and efficient pre-trained language models and tools. However, English summarization models for low-resource Indian languages are often limited by rich morphological variation, syntax, and semantic differences. 
% Also, the restricted form of supervision limits the generality and usability of low-resource languages due to the lack of annotated corpora. 
In this paper, we propose \textsc{GAE-ISumm}, an unsupervised Indic summarization model that extracts summaries from text documents.
In particular, our proposed model, \textsc{GAE-ISumm} uses Graph Autoencoder (GAE) to learn text representations and a document summary jointly. 
% Positional and cluster information is also embedded\textcolor{red}{incorporated} while extracting the summary of the document. 
% and leverages: (i) learning document representations and (ii) jointly learning sentence representations and summary of the document.
We also provide a manually-annotated Telugu summarization dataset \textsc{TELSUM}, to experiment with our model \textsc{GAE-ISumm}. 
Further, we experiment with the most publicly available Indian language summarization datasets to investigate the effectiveness of \textsc{GAE-ISumm} on other Indian languages. 
% Further, we investigate the effectiveness of \textsc{GAE-ISumm} on 
% \textsc{TELSUM}, and as well as in seven Indian languages from existing datasets.
% Recently, graph autoencoder (GAE) has shown superior performance on several NLP tasks even with limited resources.
%To evaluate our model, we also introduce a manually annotated Telugu Summarization dataset (\textsc{TELSUM}) of 501 document-summary pairs.  \textcolor{red}{why did we introduce it for Telugu alone or should we remove about dataset compeltely?}
 %\textcolor{red}{change sentence formation?}
% \textbf{Continuation for Indic from Telugu}
% To this end, we introduce a manually annotated Telugu summarization dataset (\textsc{TELSUM}) of 501 document-summary pairs to evaluate our model. 
%\textcolor{red}{We also extend our model to other Indian languages.}
% Our unsupervised model is either competitive (or) at best in comparison to the state-of-art results on all datasets. 
Our experiments of \textsc{GAE-ISumm} in seven languages make the following observations: (i) it is competitive or better than state-of-the-art results on all datasets, (ii) it reports benchmark results on \textsc{TELSUM}, and
(iii) the inclusion of positional and cluster information in the proposed model improved the performance of summaries. 
% \textcolor{red}{change the last statement.(this one}
We open-source our dataset and code~\footnote{\url{https://github.com/scsmuhio/Summarization}}.
\end{abstract}


\section{Introduction}
\label{sec:intro}
%About summarization and types of summarization(mention about pos score, sentence length.. )
Document summarization aims to minimize the content in a text document and preserve the salient information.
There are usually two categories of summarization techniques: Extractive~\cite{nallapati2017summarunner} and Abstractive~\cite{paulus2017deep}. Extractive summarization extracts salient text from the document.
Whereas Abstractive summarization concisely paraphrases the information contained in the document. 
Understanding text's contextual and semantic representation is one of the main challenges for effective summaries. Traditional methods extract summaries based on hand-crafted features, including Term Frequency~\cite{luhn1958automatic}, Sentence Position, and Length~\cite{erkan2004lexrank}, and largely depend on the availability of NLP tools.

% Traditional methods for extractive summarization can be broadly classified into greedy approaches (e.g., (Carbonell and Goldstein
% 1998)), graph based approaches (e.g., (Radev and Erkan
% 2004)) and constraint optimization based approaches (e.g.,
% (McDonald 2007)).

%Few works have studied the effectiveness of extractive summarization .
%Single-document summarization is one of the challenging problems in Information Retrieval and Natural Language Processing (NLP).  
% from the source. Abstractive summarization paraphrases the salient information identified in the document and outputs the summary. 

%Neural models, where we talk about GAE and GCN starting with Extractive summarization
%Literature survey manifests that a vast amount of work has been devoted to extractive summarization in NLP. 
Significant progress has been made in single document extractive summarization by using recent popular deep learning models such as RNNs~\cite{nallapati2017summarunner}, CNNs~\cite{zhang2016extractive}, attention-based models~\cite{ren2017leveraging}, and sequence to sequence models~\cite{nallapati2016abstractive}. 
%for English~\cite{nallapati2017summarunner}.Literature survey manifests that majority of the extractive summarization techniques rely on learning efficient text representations by using 
%Deep neural models, such as Recurrent Neural Networks (RNN)~\cite{nallapati2017summarunner}, Convolutional neural networks (CNN)~\cite{zhang2016extractive} and attention-based models~\cite{ren2017leveraging} have been successful in extractive summarization with their strong representation power. 
%Several researchers also apply the idea of RNN and attention to the abstractive summarization models~\cite{paulus2018deep,chowdhury2020neural}.
% ~\cite{hanunggul2019impact}
% ~\cite{khatri2018abstractive}
Also, most of these models follow encoder-decoder-based approaches to generate summaries in a supervised or unsupervised setting~\cite{xu2020unsupervised}.
% proposes a hierarchical transformer model to extract summaries.
% Similarly, ~\cite{liu2019text} introduces a document-level encoder using transformers to generate summaries, minimizing the reconstruction loss. 
%These models have been developed either in a supervised or. 
However, these models lack in capturing the global context and the long-distance sentence relationships present in a document. 
Modeling long-range inter-sentence relationships with transformer-based models are still challenging~\cite{xu2019discourse} and require massive computation and memory. 
By capturing the long-term dependencies and treating the document as a graph\cite{kipf2016semi}. , graph-based approaches assist in overcoming these limitations.

Recently, Graph based models have been used to capture the cross-sentence relationships for summarization~\cite{cui2020enhancing, xu2020discourse}. They have also proven their dominance in other fields such as classification~\cite{yao2019graph,marreddy2022multi}, and semantic role labelling~\cite{marcheggiani2017encoding}.
% Recently, Graph Convolutional Networks (GCN) has been successful in NLP and applied to various tasks such as text classification~\cite{yao2019graph}, semantic role labelling~\cite{marcheggiani2017encoding} and summarization~\cite{xu2020discourse}. 
Graph-based models are capable of
drawing syntactic information, exploiting long-range multi-word relations, and have been deployed on document-word relationships~\cite{liu2019text, hermann2015teaching}.
%GCN-based models adopt building a single large graph on the corpus that manages to learn reduced dimension representations and preserve the graph’s global structural information.
% \textcolor{red}{can we mention a single large graph on the corpus, because we are building sentence level also}
% \cite{xu2020selective} proposes a Graph-based selective attention mechanism preserving the syntactic and semantic structure of the document achieving state-of-art on CNN/Daily Mail dataset \cite{hermann2015teaching}. 

\begin{figure*}[!htb]
  \centering
  \includegraphics[width=\linewidth]{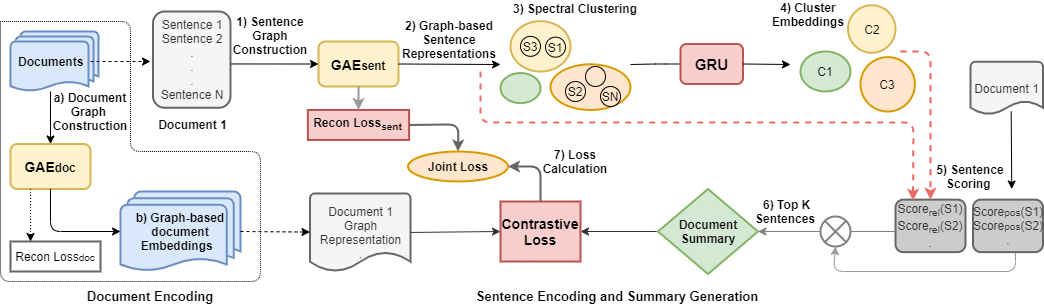}
  \caption{ Outline of \textsc{GAE-ISumm}. Our model \textsc{GAE-ISumm} involves two phases: Document Encoding - a) Document Graph Construction and b) Obtaining Graph-based Representations;  Sentence Embedding and Summary Generation involves seven steps starting from 1) Sentence Graph Construction to 7) Loss calculation.}
  \label{fig:proposedModel}
\end{figure*}

\begin{figure}[t]
\centering
    \includegraphics[width=\linewidth]{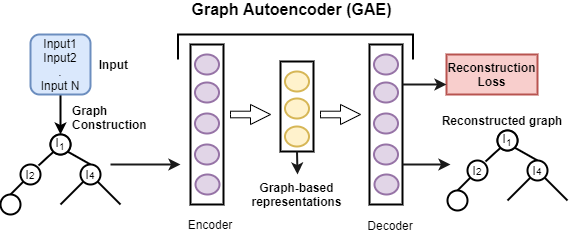}
    \caption{ GAE model}
    \label{fig:gae model}
\end{figure}

\begin{figure}[t]
\centering
    \includegraphics[width=\linewidth]{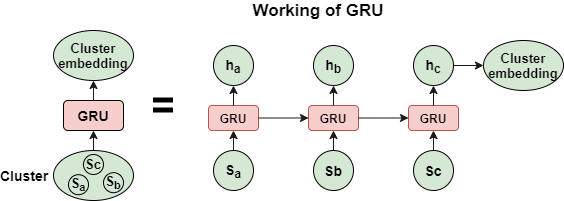}
    \caption{ Working of GRU }
    \label{fig:grumodel}
\end{figure}

%\textcolor{red}{connection missing?}
To make our model more effective and computation friendly, we get to the idea of a recently proposed method, Graph Autoencoder (GAE)~\cite{kipf2016variational}. 
GAE here captures the hidden semantic information between documents and sentences by using the idea of an autoencoder (AE)~\cite{schmidhuber2015deep} to graph-structured data.
Few recent papers have also obtained benchmark results on text classification using GAEs~\cite{xie2021inductive} . 
However, the application of GAE to document summarization is an unexplored area.
Also, adopting an AE or GAE trained on English corpora for Indian languages have significant limitations: (i) lack of such large-scale human-annotated datasets, (ii) rich morphological variation, and (iii) syntactic and semantic differences. 
%\textcolor{red}{can we say this now, nona-availability of pre-trained models}
%is often limited by data availability, rich morphological variation, syntax, and semantic differences.
%In addition, the resource-poor languages like Telugu have significant limitations: (i) lack such large-scale human-annotated datasets, (ii) non-availability of efficient pre-trained models.
%Due to the these limitations, summarization models are not well studied for low resource Indian languages. 
Moreover, due to the dearth of various qualitative tools and scarcity of annotated data, summarization models are not well studied for low-resource Indian languages.
Unlike English, few preparatory works have studied text summarization for Indian languages.
In these works, the authors use existing methods such as keyword extraction~\cite{naidu2018text}, hand-crafted features~\cite{gulati2017novel}, and 
TextRank algorithm~\cite{manjari2020extractive} to extract the summaries.
In another recent work, ~\cite{mamidala2021heuristic} proposes a heuristic model based on the frequency score of named entities and the vocabulary of the document to produce summaries.
% In another recent work, ~\cite{joshi2021sgats} constructs a semantic graph on text documents by establishing semantic relationships with the help of Wordnet. 
% graph of the original Hindi text document by establishing a semantic relationship between sentences of the document using Hindi Wordnet ontology as a background knowledge source.
Unfortunately, all these works were limited to existing baseline models that rank sentences based on term frequency or similarity heuristics. Also, none of the works focused on learning sentence or document representations, which is crucial for an effective summarization model. 
%In addition, resource-poor languages like Telugu have significant limitations: (i) lack such large-scale human-annotated datasets, and (ii) non-availability of efficient pre-trained models.
In this scenario, unsupervised approaches are alluring as they do not require any labeled data for training.

Inspired by the GAE~\cite{kipf2016variational}, SummPip ~\cite{zhao2020summpip}, and Salience Score Estimation~\cite{yasunaga2017graph}, we propose \textsc{GAE-ISumm}:  an unsupervised extractive summarization model that jointly optimizes the loss between document representations, sentence representations and generated summaries to showcase better performance.
More importantly, our proposed model simultaneously learns the sentence representations and summary of the document, while earlier methods extract summaries after learning the sentence representations.
Fig.~\ref{fig:proposedModel} illustrates our proposed method,~\textsc{GAE-ISumm}.
% We didn't add about \textcolor{blue}{position score} in introduction.

% \textcolor{red}{New paragraph}
% We also experiment our model effectiveness on other available summarization Datasets. And what if they ask, what is the morphological element that you introduced for Indian languages.

%for the Telugu language. Our proposed method  
%In this paper, we propose \textsc{\textsc{GAE-ISumm}}: an unsupervised GAE-based extractive summarization model for the Telugu language. 
%Most of the summarization techniques first learn the sentence representations and then select sentences, whereas we propose a model that simultaneously learns the sentence representations and summary of the document. 
%\textcolor{red}{we didn't mention about the crosslingual experiments in introduction, clusterisation, should we do so?}
Our main contributions are as follows: 
\begin{itemize}
    \item We propose \textsc{GAE-ISumm}, an unsupervised model that learns a summary from the input document. Also, to the best of our knowledge, we are the first to apply GAE for the summarization task.
    \item We formulate the problem as a graph network to learn sentence and document representations using GAE and perform text summarization jointly with our proposed method. 
    \item To the best of our knowledge, we are the first to investigate the effectiveness of graph-based embeddings for different Indian languages in an unsupervised setting. %~\textcolor{red}{Is this sentence fine?}
    \item We further introduce \textsc{TELSUM}, a manually annotated Telugu summarization dataset of 501 document-summary pairs.
    \item We experimented our proposed model, \textsc{GAE-ISumm}, on different datasets with monolingual and bilingual settings across seven Indian languages.
\end{itemize}

This paper aims to bridge the gap by creating models and resources for the summarization task of Indian languages. The proposed method, \textsc{GAE-ISumm}, can be extended to other resource languages that are closer to Indian languages culturally and linguistically by translating this resource without losing the rich morphological variations.

\section{Datasets}

Most of the summarization datasets of Indian languages are web scraped and consider the headline or the first lines of the text as summary~\cite{hasan2021xl}. 
Table~\ref{tab:all datasets statistics} reports the statistics of different summarization datasets available for Indian languages.
% From Table~\ref{tab:all datasets statistics},
Based on the compression ratio, we notice that the summary of the article is either under-representing or over-representing the document.
% the summary length is either too long or too short compared with the document size for many Indian language datasets.
%This indicates that the summary is either under-representing or over-representing the document.
Unlike existing datasets, we introduce a dataset \textsc{TELSUM} for Telugu that is manually annotated (human-written summaries) to create a gold-standard summarization dataset. Here we provide a brief overview of our TELSUM dataset and other existing Indic datasets, which we employ in our experiments.

% Here, we briefly discuss about our TELSUM dataset and other existing Indic datasets that are used for our experiments.\textcolor{red}{change wording} \textcolor{green}{Here we provide a brief overview of our TELSUM dataset along with other Indic datasets already available on which we employ in our experiments.}
%We experiment with our proposed model \textsc{GAE-ISumm} on different existing summarization datasets. 
% The overall statistics of \textsc{TELSUM} and other datasets is presented in Table~\ref{tab:all datasets statistics}. 
% There are a few other available datasets XL-SUM, Hindi Short Summarization Dataset, Marathi Summarization Dataset.
% We describe each of the dataset below. 

\subsection*{\textbf{\textsc{TELSUM}: New Telugu Summarization Dataset}}

%Here, we describe the dataset collection process, preprocessing steps, and annotation details.
% To build an annotated dataset, 
For \textsc{TELSUM}, we scraped a total of 4020 documents (news articles) from a Telugu news website~\textit{samyam}\footnote{\url{https://telugu.samayam.com/}}. 
After crawling the documents, we cleaned and preprocessed~\footnote{\url{https://tinyurl.com/2p8pf4kr}} the data by removing the unwanted URLs, hash-tags, hyperlinks, English text, and redundant documents (<5 sentences).
Post-processing, we obtained a total of 3098 documents, of which 2597 documents are used for training the \textsc{GAE-ISumm} in an unsupervised setting, and the remaining documents (501) form the test dataset, henceforth referred to as \textsc{TELSUM}\footnote{\url{https://github.com/scsmuhio/Summarization}}.
Our \textsc{TELSUM} dataset consists of 501 document-summary pairs. These summaries were manually written by two professional annotators who are native Telugu speakers.
Three highly proficient Telugu native speakers verified the written summaries regarding readability, relevance, and creativity. Further, we use these annotated summaries to evaluate our model \textsc{GAE-ISumm}.
The privacy details, fair compensation for annotators, and ethical concerns are discussed in Section~\ref{ethical_statement}.
Further details regarding the annotation guidelines are provided in the appendix. %\textcolor{red}{check if this is how we refer to appendix}

% \textbf{Annotation Details}
% Each of the annotator was given a set of guidelines and asked to first write sample summaries. The summaries written were verified and then asked to perform annotation on the whole dataset TELSUM. 
% Set of Guidelines: The annotator has to read the article few times, understand the whole context and information present. 
% The annotators shouldn't include bias in any form in the summary, from their perspective.
% 

\begin{table}[t]
\vspace{0.2em}
\scriptsize
\centering
\caption{ Statistics of \textsc{TELSUM} and different available summarization datasets in Indian languages. Here, the Compression ratio is the average length of the summary to the average document length.}
\label{tab:all datasets statistics}
\resizebox{0.4\textwidth}{!}
{\begin{tabular}{|p{1.2cm}|p{1.3cm}|p{1cm}|p{1cm}| p{1.1cm}|}
% {\begin{tabular}{|l|c c c c c|}
\hline
% \multicolumn{1}{|c|}{} & \multicolumn{3}{c|}{\textbf{Fine-Tuning}} \\ 
% \cline{1-6} 

Dataset & Lang &\#Docs & Avg len of doc & Compression ratio \\ \hline 
 
TELSUM & Telugu (te) & 501 & 17.43 & 0.170 \\
NCTB & Bengali (bn) & 200 & 08.06 & 0.518 \\
BNLPC & Bengali (bn) & 139 & 12.78 & 0.129 \\
Marathi  & Marathi (mr) & 100 & 14.80 & 0.513 \\
Hindi Short & Hindi (hi) & 66,367 & 17.91& 0.056  \\ \hline
& Telugu (te) & 11,308 & 31.14 & 0.032  \\
 & Bengali (bn) & 8,226 & 41.53 & 0.024 \\
 & Tamil (ta) & 17,846 & 33.59 & 0.030  \\
XL-sum & Gujarati (gu) & 9,665 & 49.68 & 0.020 \\
& Punjabi (pa) & 8,678 & 41.02 & 0.020 \\
& Marathi (mr) & 11,164 & 55.13 & 0.018 \\
& Hindi (hi) &  51,715 & 29.79 & 0.036 \\

\hline
%\multicolumn{4}{c}{ P = Precision, R = Recall, F1 = F1-score} \\
\end{tabular}}

\end{table} 

\subsubsection*{Other Datasets}

\noindent\textit{XL-sum:}
XL-sum is a comprehensive and diverse dataset~\cite{hasan2021xl} extracted from the BBC website. It covers 44 languages, of which we consider seven Indian languages and English for our experiments. 
% Also, we use the English subset of XL-sum for our cross-lingual experiments. The dataset uses contextual information or the gist as the summary.

\noindent\textit{NCTB and BNLPC:}
NCTB and BNLPC~\cite{chowdhury2021unsupervised} are Bengali abstractive and extractive summarization datasets extracted from Bangladesh textbooks and Bengali daily newspapers, respectively.
% NCTB~\cite{chowdhury2021unsupervised} is a Bengali abstractive dataset collected from Bangladesh textbooks. Similarly, BNLPC~\cite{chowdhury2021unsupervised} is a Bengali extractive dataset extracted from Bengali daily newspapers.
%'The Daily Prothom alo' and 'The Daily Jugantor.'

\noindent\textit{Marathi, and Hindi Short Summaries:}
Marathi dataset is extractive and collected from a news website~\footnote{\label{note4}\url{https://tinyurl.com/mtz473dr}}. 
Hindi short summarization dataset consists of 330k articles scraped from a news Hindi website~\footnote{\label{note5}\url{https://tinyurl.com/4pd86b55}} which considers the headline as the summary of the article.

\section{Methodology}
In this section, we explain (i) the overview of GAE, (ii) Graph construction, and (iii) a detailed description of the proposed method \textsc{\textsc{GAE-ISumm}}
% (iii) the procedure to build document and sentence level graphs, (iv) the GRU model for cluster embedding, and (v) the estimation of sentence salience scores and selection.

% Our model broadly involves two steps: Learning graph-based document embedding with the help of GAE doc, and learning the summary based 
% Our Model first learns the text representation with the help of GAE, then we apply clustering to obtain semantic clusters. Now, we score sentences with respect to the learned representations and the relative sentence position in the document and select top K sentences as summary. 
% The whole model is trained from end-to-end using a joint loss function~\eqref{eqn:Joint loss}.

% \vspace{-0.2cm}

\subsection{\text{Graph Autoencoder (GAE)}}
%introduction about GAE
Graph Autoencoder (GAE)~\cite{kipf2016variational} takes input, an undirected weighted graph $\mathcal{G}$ := ($\mathcal{V}$, \textit{A}, \textit{X}), where $\mathcal{V}$ is a set of N nodes $(v_1,v_2,..v_N)$, \textit{A} $\in\mathbb{R}^{NXN}$ is a symmetric adjacency matrix representing node relationships and \textit{X} $\in\mathbb{R}^{NXD}$ is the node feature matrix. GAE obtains an encoding \textit{Z} $\in\mathbb{R}^{NXP}$, a reduced dimension space of \textit{X}. The GAE model tries to reconstruct an Adjacency matrix \textit{A}$^{'}$ close to the empirical graph (\textit{A}) while \textit{Z} captures the essential components of $\mathcal{G}$. We stack an inner product decoder to reconstruct the graph as follows:
\setlength{\belowdisplayskip}{0pt} \setlength{\belowdisplayshortskip}{0pt}
\setlength{\abovedisplayskip}{0pt} \setlength{\abovedisplayshortskip}{0pt}
{\small
\begin{align}
   \label{gae}
   A^{'} = g(AH^{(1)}\Theta_{1}) \\  
    Z = H^{(1)} = f(AX\Theta_{0})
\end{align}
}%
where \textit{g} is an activation function, and $\Theta_0$, $\Theta_1$ are the weights learned from the graph reconstruction. Fig.~\ref{fig:gae model} briefly describes the workflow of the GAE model.
%The above equations help us to understand the working of GAE.

\subsection{\text{Graph Construction}}
The \textsc{GAE-ISumm} incorporates graph construction at two levels - one at the sentence level (each node representing a sentence in the sentence graph) and another at the document level (each node representing a document in the document graph), as shown in Fig.~\ref{fig:proposedModel}.
%We build two text graphs either at the sentence level (each node representing a sentence) (or) a document level (each node representing a document).
We use cosine similarity to identify the relationship between the nodes, which helps us build the graph's adjacency matrix. The goal of graph construction is to capture the global context either within the document (sentence level graph) (or) across all the documents (document level graph).

\subsection{\textsc{\textsc{GAE-ISumm}}}
The overall pipeline of our proposed model \textsc{\textsc{GAE-ISumm}} is described in Fig.~\ref{fig:proposedModel}. Our proposed model involves training in two phases: i) document encoding and ii) sentence encoding and summary generation, where each phase is trained separately.
The following subsections explain all the key components of \textsc{GAE-ISumm}.

\subsubsection{Document Level Graph Construction and Encoding:}
% \noindent\textbf{Document Level Graph Construction and Encoding:}
\label{Document Level Graph Construction and Encoding}
%In document level graph, each document $D_j$ in the whole dataset $(D_1,D_2,...)$ acts as a node. 
To build a document-level graph across all the documents, first, we obtain sentence representations for each document as further discussed in section \nameref{Sentence Encoding}
% using available  pre-trained language models 
% % such as Word2Vec, FastText, BERT, and ALBERT
% ~\cite{devlin2018bert,conneau2019unsupervised, xue2020mt5,kakwani2020indicnlpsuite, marreddy2021clickbait,marreddy2022resource}  as described in sentence encoding and summary generation.
% sentence-level graph construction and encoding.
In order to get the document representation, we map each node or document $D_j$ to a fixed-length vector ($X_{doc}$) by averaging all the sentence representations in $D_j$.
% Each document $D_i$ (or) node is mapped to a fixed-length vector using the pre-trained language models \cite{marreddy2021clickbait}. First, We obtain sentence mappings as described in ~\ref{Sentence Encoding and summary generation}. 
% Then, each document representation is obtained by averaging all the sentence representations (from the pre-trained model) in $D_i$. We obtain sentence representations as described in ~\ref{Sentence Encoding and summary generation}. 
Finally, the document-level graph is fed into GAE$_{doc}$ to obtain the graph-based latent document representations ($Z_{doc}$). 
The GAE$_{doc}$ model is trained independently by minimizing the reconstruction loss of document level graph (${\text{Recon\,loss}}_{doc}$). 
These obtained latent document representations ($Z_{doc}$) are further used in the \textsc{GAE-ISumm} model while calculating loss. 
%\textcolor{red}{does the last sentence seem fine?}
% at the end to learn summaries. 
% ${GAE}_{doc}$ helps in learning representations at a global level.

% \begin{table}[t]
% \vspace{0.2em}
% \scriptsize
% \centering
% \resizebox{0.5\textwidth}{!}
% {\begin{tabular}{|l|p{1cm}|p{1cm}|p{1cm}|}
% % {\begin{tabular}{|l|c c c c c|}
% \hline
% % \multicolumn{1}{|c|}{} & \multicolumn{3}{c|}{\textbf{Fine-Tuning}} \\ 
% % \cline{1-6} 

% \textbf{Language} &\textbf{No. of Documents} & \textbf{Avg len of document} & \textbf{Compression ratio }      \\ \hline 
 
% Telugu & 11,308 & 31.14 & 0.032  \\
% Bengali & 8,226 & 41.53 & 0.024 \\
% Tamil & 17,846 & 33.591 & 0.0297  \\
% Gujarati & 9,665 & 49.68 & 0.0201 \\
% Punjabi & 8,678 & 41.02 & 0.0204 \\
% Marathi & 11,164 & 55.13 & 0.018 \\
% Hindi & 51, 715 & 29.79 & 0.0356 \\
 
% \hline
% %\multicolumn{4}{c}{ P = Precision, R = Recall, F1 = F1-score} \\
% \end{tabular}}
% \caption{ Statistics of Different languages from XL-SUM dataset}
% \label{tab:XLSUM statistics}

% \end{table} 

\subsubsection{Sentence Encoding and Summary generation}
\label{Sentence Encoding and summary generation}

Here, we try to explain the remaining components of our model in detail. They are organized in the following manner: (i) sentence level graph construction, (ii) sentence encoding, (iii) clustering and cluster embeddings, (iv) sentence scoring and selection, and (v) loss calculation.
All these components are processed at a single document level.
%\textcolor{red}{we divided the first section into two parts}

\subsubsection*{1) Sentence Level Graph Construction} 
We build a sentence-level graph for each document D := $(S_1, S_2,..,S_N)$ where each sentence $S_i$ is considered as a node. 
To obtain a sentence-level graph, we use existing Indic pre-trained language models~\cite{kakwani2020indicnlpsuite,marreddy2021clickbait,marreddy2022resource} and various multilingual pre-trained models~\cite{devlin2018bert, conneau2019unsupervised,xue2020mt5} to get a fixed-length vector representation for each sentence. 
The pre-trained language models help us detect the high-level contextual features capturing precise semantic and syntactic relationships.
We also investigate monolingual distributed word embeddings and pre-trained language models available for Telugu language~\cite{marreddy2022resource}. After mapping each node to a fixed-length vector, the sentence-level graph is fed into ${GAE}_{sent}$ to obtain latent graph-based representations ($Z_{sent}$) of each sentence in the document.

%
%\textcolor{red}{add continuation to the embeddings part}
%\textcolor{red}{also differentiate between embeddings description and side heading}
%In addition, we use the distributed word representations like Word2Vec~\cite{mikolov2013distributed},  FastText~\cite{joulin2017bag} and context level features like BERT~\cite{kenton2019bert} and ALBERT~\cite{lan2019albert} to obtain the sentence representations $X_{sent}$. \textcolor{red}{this sentence ok?} 

\subsubsection*{2) Sentence Encoding}
\label{Sentence Encoding}

We experiment with different pre-trained language models to obtain text representations. The details of the language models are as follows:
% The details of distributed word embeddings and pre-trained language models used for sentence encoding are discussed below: \textcolor{red}{check if this statement is needed or not.}
% \noindent\textbf{Word2vec-Te, Glove-Te, and FastText-Te}: 

\noindent\textit{Distributed Word Embeddings:}
We use Word2vec-Te, Glove-Te, and FastText-Te word embedding models, trained on a large Telugu dataset of 8 million sentences~\cite{marreddy2022resource}. We average the word embeddings in the sentence to obtain the sentence representation.
%\textcolor{red}{change this sentence}

\noindent\textit{Pre-trained Telugu Transformer Language Models}:
We use the monolingual pre-trained Transformer models such as BERT-Te, Albert-Te, and Roberta-Te available for Telugu~\cite{marreddy2022resource}. 

\noindent\textit{Multilingual Embeddings}
Recently, the NLP community has contributed large-scale multilingual models for Indian languages performing exceptionally well compared to other representations.
% These language models are showing exceptional results compared to the other text representations. \textcolor{red}{cite something here and change wording.} 
We experiment with few such multilingual models: IndicBERT~\cite{kakwani2020indicnlpsuite}, mBERT~\cite{devlin2018bert}, XLM-R~\cite{conneau2019unsupervised}, mT5~\cite{xue2020mt5} to obtain sentence representations.

% \noindent\textbf{IndicBERT}: IndicBERT~\cite{kakwani2020indicnlpsuite} is a multilingual pre-trained Transformer language model created for twelve Indian languages. It is trained on an Indic-Corp~\cite{kakwani2020indicnlpsuite}. We use these IndicBERT embeddings to obtain the sentence-level representations. Here, the sentence representation is obtained by taking the average of the token representations from the last hidden state.

% \noindent\textbf{mBERT}: mBERT~\cite{devlin2018bert} is a multilingual version of BERT trained on 104 languages from Wikipedia.

% \noindent\textbf{XLM-R}: XLM-R~\cite{conneau2019unsupervised} is a transformer-based multilingual masked language model pre-trained on Common Crawl~\cite{conneau2019unsupervised} data in 100 languages. 
%It is trained only with an objective of the Masked language Model(MLM) in a RoBERTa way.

% \noindent\textbf{mT5}: mT5~\cite{xue2020mt5} is a multilingual variant of T5 model pre-trained on mC4 dataset~\cite{xue2020mt5} which contains 101 different languages.
%\textcolor{red}{I cited here twice, should I cite or should I not?}

% \noindent \textbf{Graph based Sentence Representations:}

\subsubsection*{3) Clustering and Cluster Embeddings} 
A document usually consists of multiple events (or) a series of events; we believe that clustering on the document helps to segregate better and understand the document. To accomplish this, we generate cluster representations to incorporate the cluster information in the final summary. We applied spectral clustering~\cite{ng2002spectral} on the latent sentence representations $Z_{sent}$ obtained from ${GAE}_{sent}$, where the spectral clustering method partitions a document of N sentences into M clusters $(c_1,c_2,.,c_M)$.

%Clustering helps to have an overview of the whole document and avoids redundancy in the document summary.\textcolor{red}{needed?}
% {Cluster Embedding Generation:} 
%We apply a feedforward Gated Recurrent Unit (GRU)~\cite{cho2014learning} on each cluster to obtain the cluster embedding (see Figure~\ref{Fig:proposed model}). 

For each cluster $c_i$ with $|c_i|$ sentences, the GRU~\cite{cho2014learning} mechanism outputs a cluster embedding $C_i$ on top of sentence embeddings in cluster $c_i$ (please refer Fig.~\ref{fig:grumodel}). Here, the sentences are passed into GRU according to their relative position in the document. Finally, we extract the last hidden state $h_{|c_i|}$ of GRU to obtain cluster embedding $C_i$, as shown in Equation~\eqref{eqn:hidden state}. This cluster embedding has a semantic overview of the entire cluster, which helps to capture significant text.

% Cluster embedding understands the entire cluster, which helps to capture the most relevant sentence in the cluster.
\setlength{\belowdisplayskip}{0pt} \setlength{\belowdisplayshortskip}{0pt}
\setlength{\abovedisplayskip}{0pt} \setlength{\abovedisplayshortskip}{0pt}
{\small
\begin{align}
   \label{eqn:cluster embedding}
  & h_t = \text{GRU}(h_{t-1}, {S_t}^i)\\  
    \label{eqn:hidden state}
  & C_i = h_{|c_i|}
\end{align} 
}%
\noindent where ${S_t}^i$ represents sentence at $t^{th}$ time unit in cluster $c_i$.
% Inspired by \cite{yasunaga2017graph}, we apply GRU  on each of the semantic clusters to obtain a 
% Cluster embedding help to capture the most relevant sentence of all sentences in the cluster.
% Cluster embedding helps to have an understanding of the entire cluster.
% The clusters represent different subevents that are part of the document. 
% In spectral clustering, we obtain the Laplacian matrix of the sentence level graph and compute the first K eigenvectors of the matrix to represent each sentence in a feature space. Now, we run K-Means clustering algorithm on these features dividing them into k groups $(C_1,C_2,...,C_K)$.
% \subsubsection{Cluster Embedding}

\subsubsection*{ 4) Sentence Scoring and Selection}
For each sentence $S_i$ of cluster $c_j$ in the document D, we estimate the sentence score using two criteria, (i) Sentence relevance score $({score}_{rel})$ and (ii) Sentence position score $({score}_{pos})$. 
% sentence relevance score
To estimate ${score}_{rel}$, we first calculate weighted relevance scores using the $f(S_i, D)$ in Equation~\eqref{eqn:rel score first} similar to attention mechanism~\cite{vaswani2017attention}.
Later, the scores are normalized via Softmax to obtain ${score}_{rel}$ as shown in Equation~\eqref{eqn:relevance score second}.
% \begin{equation}
%     \label{eqn:rel score first}
%     f(S_i, D) = w^T \tanh(W_1*S_i + W_2*C_j)
% \end{equation}
% \begin{equation}
%     \label{eqn:relevance score second}
%     {score}_{rel}(S_i, D) = \frac{f(S_i)}{ {\Sigma}_{ {S_p}\in {C_j} } f(S_p) }
% \end{equation}
% \setlength{\belowdisplayskip}{0pt} \setlength{\belowdisplayshortskip}{0pt}
% \setlength{\abovedisplayskip}{0pt} \setlength{\abovedisplayshortskip}{0pt}

{\small
\begin{align}
   \label{eqn:rel score first}
   & f(S_i, D) = \omega^T \tanh(W_1*Z_{sent}^{S_i} + W_2*C_j)\\  
    \label{eqn:relevance score second}
   & {score}_{rel}(S_i, D) = \frac{f(S_i,D)}{ {\Sigma}_{ {S_p}\in {C_j} } f(S_p,D) }
\end{align}
}%
\noindent where $Z_{sent}^{S_i}$ denotes the graph-based latent sentence representation of $S_i$, $C_j$ represents its cluster embedding, and \{$\omega$, $W1$, $W2$\} are trainable parameters. Inspired from~\cite{joshi2019summcoder}, ${score}_{pos}$ (refer Equation ~\eqref{eqn:pos_score}) is calculated based on the relative position $P(S_i)\in $ [1,2,..,N] of sentence $S_i$ in the Document D := $(S_1, S_2,..,S_N)$. Sentences at the start of the document are given high priority than the rest as they provide more relevant information about the entire document~\cite{lin1997identifying}. The final sentence score is calculated as shown in Equation~\eqref{eqn:final_score}.

\setlength{\belowdisplayskip}{0pt} \setlength{\belowdisplayshortskip}{0pt}
\setlength{\abovedisplayskip}{0pt} \setlength{\abovedisplayshortskip}{0pt}
{\small
\begin{equation}
    \label{eqn:pos_score}
    {score}_{pos}(S_i, D) =  \max\left(0.5, exp\frac{-P(S_i)}{\sqrt[3]{N}} \right)
\end{equation}
}%
\setlength{\belowdisplayskip}{0pt} \setlength{\belowdisplayshortskip}{0pt}
\setlength{\abovedisplayskip}{0pt} \setlength{\abovedisplayshortskip}{0pt}
{\small
\begin{equation}
\label{eqn:final_score}
    \text{score}(S_i,D)=\alpha*{\text{score}}_{rel}(S_i,D) + \beta* {\text{score}}_{pos}(S_i,D)
\end{equation}
}%
%\normal
The variables in Equation~\eqref{eqn:final_score}: $\alpha$, $\beta$ $\in$ [0,1] with $\alpha$+$\beta$=1,  assign relative weights to ${score}_{rel}$ and ${score}_{pos}$, respectively.
% \textcolor{red}{long sentence?}
In every iteration, we sort the sentences in descending order of their sentence scores, and the top K sentences are considered our predicted summary ($\hat{S}$) of the document.
We obtain our final candidate summary whenever our model reaches the local minima solution. %\textcolor{red}{should we mention about convergence?}

\subsubsection*{5) Loss calculation} 
We estimate the contrastive loss \cite{hadsell2006dimensionality} between the graph-based latent document representations ($Z_{doc}^D$) and the average of all sentence representations in the candidate summary. The final loss $\mathcal{L}$ is calculated as follows:
%\textcolor{red}{add balancing factors?}

{\small
\begin{align}
\label{eqn:Joint loss}
\mathcal{L} &= \text{Reconstruction loss}({GAE}_{sent}) + \\ \nonumber
 & \text{Contrastive loss}(\hat{S}, Z_{doc}^D)
\end{align}
}%
%where $\hat{S}$ denotes the predicted summary.
% \textbf{Loss function and Training: }We consider the graph-based representation of each document as our gold summary. And calculate the 
% Contrastive loss\cite{hadsell2006dimensionality} between the gold summary and the candidate summary. We train the model to minimize the joint loss L ~\eqref{eqn:Joint loss}, summation of reconstruction loss of ${GAE}_{sent}$ and contrastive loss over all the documents in training data. 

%\setlength{\tabcolsep}{1pt}

\section{Experiments and Evaluation}
This section describes \textsc{GAE-ISumm} training setup, hyper-parameter tuning, and evaluation metrics.

\subsection{Model Training Setup \& Hyperparameters}
% \vspace{-0.1cm}
% \textcolor{red}{include results on English datasets. also, if required, move a few of the results to the appendix. }
\label{Training and Evaluation}
Here, we describe the training details of document encoding (${GAE}_{doc}$), sentence encoding (${GAE}_{sent}$), and summary generation over all documents.
In our \textsc{GAE-ISumm}, we train each phase (${GAE}_{doc}$, ${GAE}_{sent}$ and summary generation) separately. The final training was performed by minimizing the joint loss (\text{Reconstruction loss} + \text{Contrastive loss}) over all the documents, as mentioned in Equation~\eqref{eqn:Joint loss}.

The model trainable parameters in ${GAE}_{doc}$, ${GAE}_{sent}$, GRU, and sentence scoring \{$\omega$, $W1$, $W2$\} are described below. 
To perform document summarization using~\textsc{\textsc{GAE-ISumm}}, we set the first convolution layer's embedding size as 128 for distributed word embeddings and  256 for remaining representations.
% for both $GAE\_doc$ and $GAE\_sent$.
% The input feature vector for the summarization task is extracted from Telugu pre-trained embeddings (Word2Vec-Te, FastText-Te, Glove-Te, BERT-Te, RoBERTa-Te, ALBERT-Te) as well as from multilingual language models (mBERT, IndicBERT, XLM-R). 
The input feature vectors are extracted from different language models with an input size of 300 for word embeddings and 768 for other representations.
% for the summarization task is extracted from Telugu pre-trained embeddings \textcolor{green}{check wording}(Word2Vec-Te, FastText-Te, Glove-Te, BERT-Te, RoBERTa-Te, ALBERT-Te) as well as from multilingual language models (mBERT, IndicBERT, XLM-R). 
Since IndicBERT~\cite{kakwani2020indicnlpsuite} shows superior performance over other multilingual models, in this paper, we use IndicBERT for input feature extraction for all other experiments. 

%\noindent 
%For document encoding, we set the edge detection threshold in the document-level graph as 0.85 for BERT-Te, RoBERTa-Te and ALBERT-Te, 0.7 for FastText-Te, and 0.4 for Word2Vec-Te and Glove-Te. 
We use Adam optimizer with an initial learning rate of 0.001 to train $GAE_{doc}$.
%\textcolor{red}{add sizes and thresholds for all the remaining embeddings also?}
%For sentence encoding and summary generation, we set the edge detection threshold in the sentence-level graph as 0.8 for BERT-Te and ALBERT-Te, 0.4 for FastText-Te, and 0.1 for Word2Vec-Te.\textcolor{red}{add values}
We use the scikit-learn package of Spectral Clustering to perform sentence graph clusterization.
After experimenting with various values, the number of clusters is considered close to the average number of sentences in annotated summaries.
The joint loss function is optimized with an Adam optimizer and a learning rate of 0.0005. 
We experimented with a range of values to determine the choice of $\alpha$ and $\beta$. 
The model was effective when $\alpha$=0.6 and $\beta$=0.4. 
To extract the summary, we chose the value of K (number of sentences in the predicted summary) dependent on the number of clusters.
%where each sentence represents a cluster capturing the complete information of the whole document. \textcolor{red}{any change in sentence?}
We train each phase of the model to a maximum of 40 epochs.
The experiments were performed on a single V100 16GB RAM GPU machine.

\subsection{Evaluation Metrics}
% \vspace{-0.1cm}
We use ROUGE~\cite{lin2002manual} F1-metric (ROUGE-1 (R-1), ROUGE-2 (R-2), ROUGE-L (R-L) ) to evaluate our model. ROUGE-N score refers to N-grams overlap between candidate and gold summary (human reference summary). While ROUGE-L refers to the longest matching sub-sequence of candidate and gold summary.

\begin{table}[t]
% \vspace{0em}
\scriptsize
\centering
\caption{Comparison of ROUGE score results of \textsc{GAE-ISumm} with other methods on \textsc{TELSUM} dataset.}
\label{tab:TELSUM results}
\resizebox{0.35\textwidth}{!}{\begin{tabular}{|l| c| c c c|}
\hline
\cline{1-5} 
\multicolumn{2}{|l|}{Model} & R-1 & R-2 & R-L      \\ \hline 
% \textbf{Baselines} & & & \\
\multicolumn{2}{|l|}{TextRank}& 36.77 & 22.14 & 34.30   \\ 
\multicolumn{2}{|l|}{LexRank} & 38.89 & 26.65 & 36.45   \\ 
\multicolumn{2}{|l|}{SumBasic} & 34.97 & 20.24 & 33.65    \\
\multicolumn{2}{|l|}{KL Greedy}& 33.85 & 18.71 & 31.77  \\ \hline
\multirow{12}{*}{\textsc{GAE-ISumm}} & Word2Vec& 42.81 & 30.13 & 41.53   \\ 
& Glove& 42.81 & 30.13 & 41.53   \\ 
& FastText& 42.49 & 33.42 & 41.49 \\
& BERT-Te& 45.12 & 34.81 & 43.83    \\
& ALBERT-Te & 44.45 & 31.78 & 43.20     \\ 
& Robert-Te& 45.84 & 37.19 & 44.69   \\ 
% & distilbert& 43.96 & 33.99 & 42.87   \\ 

& XLM-R & 45.24 & \textbf{38.49} & 44.14   \\ 
& Indicbert& \textbf{46.29} & 36.6 & \textbf{44.9}   \\ 
& mbert& 44.12 & 34.72 & 41.95   \\ 
& mT5& 45.47 & 35.08 & 43.83   \\ 
& mbart& 44.86 & 32.9 & 42.27   \\

\hline
%\multicolumn{4}{c}{ P = Precision, R = Recall, F1 = F1-score} \\
\end{tabular}}

\end{table}

\section{Results and Analysis}
The effectiveness of our proposed model is evaluated by comparing with several existing baselines such as Textrank~\cite{mihalcea2004textrank}, LexRank~\cite{erkan2004lexrank}, SumBasic~\cite{nenkova2005impact}, and KL Greedy ~\cite{haghighi2009exploring}. 
In particular, we analyze our \textsc{GAE-ISumm} method on \textsc{TELSUM} as well as on other existing Indian language datasets.
%\textcolor{red}{need to change this whole paragraph}
%The results can be analyzed on two datasets: \textsc{TELSUM} and \textsc{XL-Sum}, as reported in Tables~\ref{tab:TELSUM results} and~\ref{tab:XLSUM results}.
We also describe the performance of each component of our proposed method in the ablation study.

\subsection{Performance on \textsc{TELSUM} Dataset}
\begin{figure}[t]
%\vspace{-1em}
\scriptsize
\centering
% \textbf{Sentence} \\ \hline 
\includegraphics[width=\linewidth]{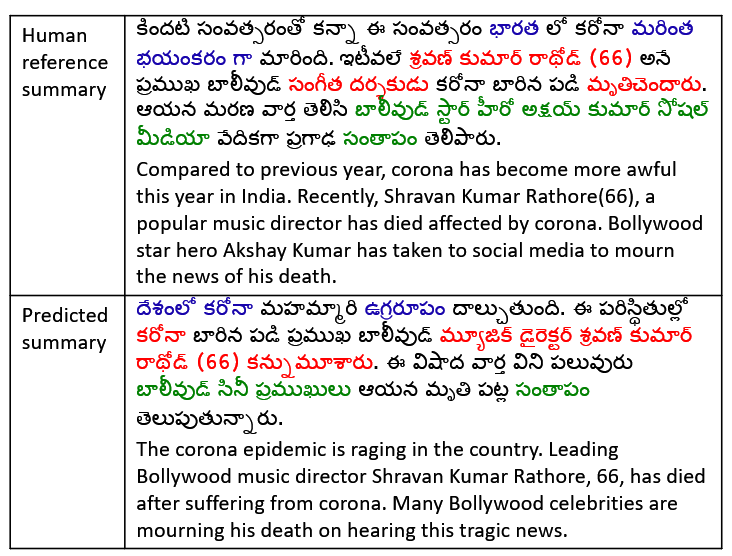}
\caption{Example of Human annotated summary and \textsc{GAE-ISumm} predicted summary from \textsc{TELSUM}}
\label{fig:TELSUM example summary}
\end{figure}

Table~\ref{tab:TELSUM results} reports the ROUGE scores of the baseline methods and our proposed method \textsc{\textsc{GAE-ISumm}} with various pre-trained language model features as input.
% features are given as input to \textsc{GAE-ISumm}
% In particular, 
We make the following observations from Table~\ref{tab:TELSUM results}:
% (i) All the baseline methods showcase lower ROUGE scores compared to \textsc{GAE-ISumm} because these methods follow simple heuristics based on sentence similarity.
(i) All the baseline models follow a basic heuristic based on similarity, which showcases lower ROGUE scores than GAE-ISumm. GAE-ISumm demonstrates its efficacy by surpassing all baselines using any Telugu or multilingual pre-trained model.
We contend that graph-based representations facilitate the capture of both global (document-level) and local (sentence-level) context information.
(ii) The LexRank model reports a higher ROUGE score among all the baseline models.
% (iii)
% \textsc{GAE-ISumm} displays its effectiveness by outperforming all the baselines with any of the Telugu or multilingual pre-trained models.
% Henceforth, we argue that graph-based representations aid in capturing both global (document-level) and local context (sentence-level) information.
% \textsc{GAE-ISumm} with all the Telugu pre-trained models outperform the baselines and display it's effectiveness.
% \textcolor{red}{check this sentence once}
% \textcolor{green}{
% All the baseline models follow basic heuristic based on phase similarity which showcases lower ROGUE scores compared to GAE-ISumm.. GAE-ISumm demonstrates its efficacy by surpassing all baselines using any Telugu or multilingual pre-trained model.
% From this point forward, we contend that graph-based representations facilitate the capture of both global (document-level) and local (sentence-level) context information.
% }
(iii) All the pre-trained transformer-based models performed similarly, while IndicBERT and XLM-R report the highest ROUGE scores for R-1, R-L, and R-2, respectively.
% Overall, \textsc{GAE-ISumm} achieve the highest ROUGE-1 score (46.29) with multilingual pre-trained language model IndicBERT~\cite{kakwani2020indicnlpsuite}. (v) All the pre-trained transformer based models 
%However, the performance is lower compared to our proposed method, \textsc{\textsc{GAE-ISumm}}.
%\noindent\textbf{\textsc{\textsc{GAE-ISumm}} Results:}
%Table~\ref{tab:TELSUM results} illustrates the performance of \textsc{\textsc{GAE-ISumm}} where various pre-trained language model features are given as input to \textsc{GAE-ISumm}. 
%We make the following observations from Table~\ref{tab:TELSUM results}: (i) \textsc{GAE-ISumm} with all the Telugu pre-trained models outperform the baselines and display its effectiveness, (ii) Overall, \textsc{GAE-ISumm} achieve the highest ROUGE-1 score (46.29) with multilingual pre-trained language model IndicBERT.
%\textcolor{red}{should we still keep XL-SUM summary example and change this sentences}
Fig.~\ref{fig:TELSUM example summary} shows human-annotated summary and the summary predicted by \textsc{GAE-ISumm} for an example article from \textsc{TELSUM} dataset.
From Fig.~\ref{fig:TELSUM example summary}, we observe that ~\textsc{GAE-ISumm} extracts all the essential named entities and highly coincide with the manual summaries in terms of coverage and consistency.
% Our model outperforms all these widely used baseline approaches. 

\begin{figure}[t]
% \vspace{-1em}
%\scriptsize
\centering
% \textbf{Sentence} \\ \hline 
\includegraphics[width=\linewidth]{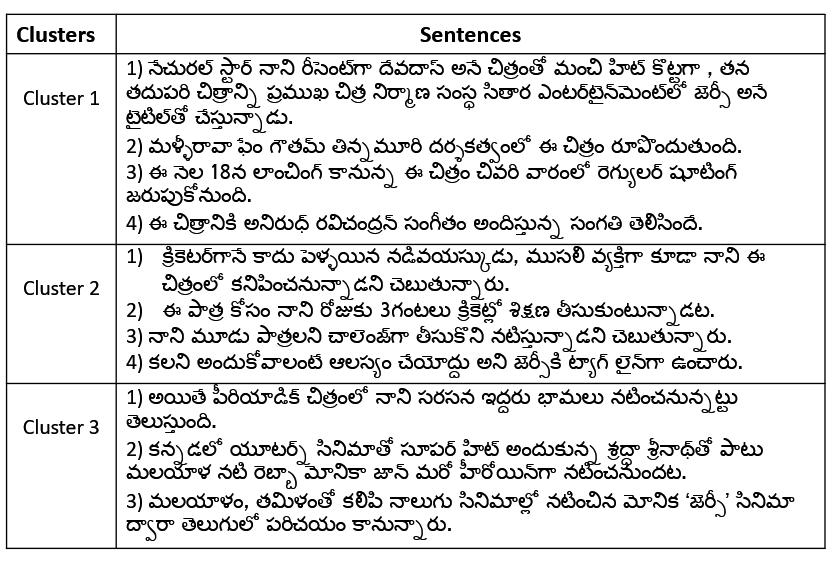}
\caption{Analysis of clusters: the clusters formed by \textsc{GAE-ISumm} for an example article from \textsc{TELSUM}.}
\label{fig: Cluster analysis}
\end{figure}

\begin{table}[t]
%\vspace{0.2em}
\scriptsize
\centering
\caption{ Ablation studies of \textsc{GAE-ISumm} on \textsc{TELSUM} dataset with Indicbert embeddings.}% \textcolor{red}{check if that line is ok in between}}
\label{tab:TELSUM Ablation studies}
\resizebox{0.4\textwidth}{!}
{\begin{tabular}{|l|c c c|}
\hline
% \multicolumn{1}{|c|}{} & \multicolumn{3}{c|}{\textbf{Fine-Tuning}} \\ 
\cline{1-4} 

Ablation studies on \textsc{GAE-ISumm} & R-1 & R-2 & R-L     \\ \hline 
 
Without ${GAE}_{sent}$ and ${GAE}_{doc}$ & 42.94& 33.04 & 39.19\\
With ${GAE}_{sent}$ and without ${GAE}_{doc}$ & 43.36 & 32.45 & 41.58 \\
With ${GAE}_{doc}$ and without ${GAE}_{sent}$ & 43.49 & 33.09 & 40.99 \\ \hline
Without clustering & 44.61 & 31.27 & 41.09\\
Without ${score}_{pos}$ & 43.14 & 32.97 & 39.95 \\ 
\hline
%\multicolumn{4}{c}{ P = Precision, R = Recall, F1 = F1-score} \\
\end{tabular}}

\end{table}

\iffalse
\begin{table*}[ht]
\begin{center}
% \vspace{0.85em}
\scriptsize
\centering
\resizebox{\textwidth}{!}
{\begin{tabular}{|l|c c c|c c c|c c c|}
\hline
% \multicolumn{1}{|c|}{} & \multicolumn{3}{c|}{\textbf{Fine-Tuning}} \\ 
% \cline{2-5} 

\textbf{Methodology} & & \textbf{Tamil} & & & \textbf{Gujarati} & & & \textbf{Bengali} &   \\ \hline 
 & ROUGE-1 & ROUGE-2&ROUGE-L&ROUGE-1 & ROUGE-2&ROUGE-L&ROUGE-1 & ROUGE-2&ROUGE-L \\ \hline
TextRank & 13.11 & 1.74 & 11.49& 12.68 & 2.97 & 10.92 & 10.71 & 2.20 & 8.46  \\ 
LexRank & 13.43 & 1.98 & 11.88 & 12.42 & 2.73 & 10.82 & 9.69 & 2.15 & 7.96   \\ 
SumBasic & 11.76 & 1.37 & 10.65 & 10.96 & 2.51 & 9.83 & 7.38 & 1.43 & 6.32   \\ 
KLSummarizer & 12.39 & 1.59 & 11.27 & 12.36 & 2.90 & 11.03 & 8.95 & 1.81 & 7.46   \\ 
LuhnSummarizer & 13.48 & 2.04 & 11.68 & 12.79 & 2.86 & 10.81 & 10.95 & 2.41 & 8.71  \\ 
\textbf{\textsc{GAE-ISumm}} & \textbf{22.35} & \textbf{10.92} & \textbf{21.96} & \textbf{19.43} & \textbf{8.79} & \textbf{16.39} & \textbf{25.27} & \textbf{13.23} & \textbf{22.64}  \\ 
XL-sum & 24.33 & 11.05 & 22.07 & 21.96 & 7.7 & 19.86 & 29.56 & 12.10 & 25.13  \\  \hline
% \textsc{GAE-ISumm} & \textbf{15.92} & \textbf{\underline{ 8.87}} & \textbf{14.25} \\ \hline

%\multicolumn{4}{c}{ P = Precision, R = Recall, F1 = F1-score} \\
\end{tabular}}
\caption{ROUGE score results on different regional language datasets of XL-sum }
\label{tab:XLSUM results}
\end{center}
\end{table*}
\fi

\begin{table*}[t]
\begin{center}
% \vspace{0.85em}
\scriptsize
\centering
\caption{\textsc{GAE-ISumm} on XL-sum: ROUGE score results on seven different Indian languages. These results are compared with mBART (MB) and IndicBART (IB) results from ~\cite{dabre2021indicbart}.}
\label{tab:XLSUM results only language}
\resizebox{0.75\textwidth}{!}
{\begin{tabular}{|l|c c c|c c c|c c c|}
\hline
% \multicolumn{1}{|c|}{} & \multicolumn{3}{c|}{\textbf{Fine-Tuning}} \\ 
% \cline{2-5} 

Metric& \multicolumn{3}{|c|}{R-1} &  \multicolumn{3}{|c|}{R-2} & \multicolumn{3}{|c|}{R-L}    \\ \hline 
Language$\downarrow$ / Method$\rightarrow$ & MB & IB&GAE-Summ&MB & IB&GAE-Summ&MB & IB&GAE-Summ \\ \hline

bn & \textbf{26.81} & 25.27 & 22.10& \textbf{10.57} & 9.55 & 9.20 & \textbf{22.45} & 21.51 & 20.10  \\ 
gu & 21.49 & 21.66 & \textbf{22.20} & 7.48 & 7.43 & \textbf{8.40} & 19.08 & 19.39 & \textbf{19.43}  \\ 
hi & \textbf{39.72} & 38.25 & 30.10 & \textbf{17.46} & 16.51 & 13.83 & \textbf{32.46} & 31.48 & 26.99   \\ 

mr & 21.46 & 22.26 & \textbf{22.75} & 9.53 & 9.94 & \textbf{9.95} & 19.26 & 20.08 & \textbf{20.14}\\ 

pa & 28.15 & \textbf{30.28} & 23.50 & 10.30 & \textbf{11.88} & 8.7 & 22.75 & \textbf{24.38} & 17.98  \\ 

te & 16.16 & 16.39 & \textbf{16.92} & 4.95 & 5.40 & \textbf{6.87} & 14.36 & 14.71 & \textbf{15.25}  \\ 

ta & 22.47 & 21.79 & \textbf{22.60} & \textbf{10.22} & 9.75 & 10.1 & 20.33 & 19.67 & \textbf{20.90}  \\  \hline
% \textsc{GAE-ISumm} & \textbf{15.92} & \textbf{\underline{ 8.87}} & \textbf{14.25} \\ \hline

%\multicolumn{4}{c}{ P = Precision, R = Recall, F1 = F1-score} \\
\end{tabular}}

\end{center}
\end{table*}

\begin{table*}[ht]
\begin{center}
% \vspace{0.85em}
\scriptsize
\centering
\caption{ROUGE score results on Other Summarization Datasets using \textsc{GAE-ISumm}. Here `-' indicates that the dataset has no state-of-art results. }
\label{tab:Other Datasets results}
\resizebox{0.85\textwidth}{!}
{\begin{tabular}{|l|c c| c c| c c|c|}
\hline
\cline{1-8} 

Metric & \multicolumn{2}{c|}{R-1} & \multicolumn{2}{c|}{R-2} & \multicolumn{2}{c|}{R-L} & state-of-art method  \\ \hline
Dataset$\downarrow$ / Method$\rightarrow$ & \textsc{GAE-ISumm} & State-of-art & \textsc{GAE-ISumm} & State-of-art & \textsc{GAE-ISumm} & State-of-art    &  \\ \hline
% \textbf{Baselines} & & & \\
BNLPC~\cite{chowdhury2021unsupervised}& \textbf{71.8} & 61.6 & \textbf{66.8} & 56.5 & \textbf{71.2} & 61.1   & BenSumm~\cite{chowdhury2021unsupervised}\\ 
NCTB~\cite{chowdhury2021unsupervised} & \textbf{12.33} & 12.17 & \textbf{3.19} & 1.92 & 10.99 & \textbf{11.35} & BenSumm~\cite{chowdhury2021unsupervised}   \\ 
Marathi \footref{note4} & \textbf{80.5} & 64.8 & \textbf{75.5}   & 59.1 & \textbf{80.0} & 66.1  & Similarity based text rank method\\
Hindi Short \footref{note5}& \textbf{20.2} & - & \textbf{12.0} & - & \textbf{16.1} & - & -\\ 
\hline
%\multicolumn{4}{c}{ P = Precision, R = Recall, F1 = F1-score} \\
\end{tabular}}

\end{center}
\end{table*}

\begin{table}[!htb]
% \begin{center}
% \vspace{0.85em}
\scriptsize
\centering
\caption{ Cross-lingual experiments of \textsc{GAE-ISumm}: Bilingual setting of ``hi'' and ``en'' with the other Indian languages from XL-sum. Here `-' represents no cross-lingual experiment for that permutation of languages.}
\label{tab:cross-lingual XLSUM results}
\resizebox{0.5\textwidth}{!}
{\begin{tabular}{|l|c c|c c|c c|}
\hline
% \multicolumn{1}{|c|}{} & \multicolumn{3}{c|}{\textbf{Fine-Tuning}} \\ 
% \cline{2-5} 

Metric & \multicolumn{2}{|c|}{R-1} &  \multicolumn{2}{|c|}{R-2} &  \multicolumn{2}{|c|}{R-L}  \\ \hline 
Lang-2$\downarrow$ / Lang-1$\rightarrow$ & hi & en & hi & en & hi & en \\ \hline
bn & \textbf{23.26} & 21.49 & \textbf{9.6}  & 9.21 & \textbf{21.45} & 19.5  \\ 
gu & \textbf{22.49} & 19.95 & \textbf{7.48} & 7.03 & \textbf{19.16} & 16.14  \\ 
hi & - & 29.56 & - & 12.78 & - & 24.07   \\ 

mr & \textbf{21.89} &  19.13 & \textbf{8.78} &  8.26 & \textbf{19.72} & 18.93\\ 

pa & 23.76 &  \textbf{24.1} & 8.68 &  \textbf{8.89} & \textbf{18.34} & 17.41  \\

te & \textbf{17.81} & 16.55 & \textbf{7.89} &  7.83 & \textbf{15.43} & 14.85  \\ 

ta & \textbf{22.36} & 21.24 & \textbf{9.79} & 9.1 & \textbf{20.1} & 19.54  \\  \hline

\end{tabular}}

% \end{center}
\end{table} 

\subsection*{Analysis of Clusters}
Fig.~\ref{fig: Cluster analysis} represents the clusters formed for an example article from~\textsc{TELSUM} dataset.(English version of these clusters is present in appendix Fig.~\ref{fig: english cluster analysis}). 
The example article is about a movie that was yet to release. 
Of the 3 clusters formed, the first cluster talks about the expected release date of the movie and the main crew involved.
The second cluster talks about the male lead and the characterization of the male lead.
The third cluster talks about the female-lead cast and her previous career details. We can see that clusterization helps us in grouping relevant information, getting the full coverage of the article, and removing redundant information present in the article.
% \textcolor{red}{can modify below sentence, this part also looks like a bit redundant.}
% From these clusters formed, we can clearly state that the inclusion of cluster formation helps us get full coverage of the article and helps us remove redundant information (sentences with similar semantics) present in the article.\textcolor{green}{We can clearly see that clusterisation helps us in grouping relevant information, get the full coverage of article and remove redundant information}
% Here, the redundant information means sentences with similar semantics.
%\textcolor{red}{can we say redundant information here?}

\subsection*{Ablation Studies}
Here, we analyze the importance of the individual components of \textsc{GAE-ISumm} and compare them with the benchmark result obtained in Table~\ref{tab:TELSUM Ablation studies}.
To investigate the performance of each component, we conducted several ablation experiments: (i) without graph components, (ii) without clustering and (iii) removal of sentence position scores.
% With the removal of $GAE_{sent}$ and $GAE_{doc}$ components; the model was not able to learn effective text representations and display lower ROUGE scores. 
In the second aspect, the model is trained without including cluster information i.e., we estimate $score_{rel}$ without any cluster information in Equation~\eqref{eqn:rel score first}.
The results reflect that, without clustering, the model fails to capture the complete significant information from the document. 
In the third aspect, we remove the sentence position score from the final sentence score estimation in Equation~\eqref{eqn:final_score}. 
We observe that removing positional information yields a relative drop of 6.8\% in R-1, proving the importance of sentence position while generating summaries.
% while generating summaries, as reported in Table~\ref{tab:TELSUM results}.

%\noindent\textcolor{red}{merged properly?}

\noindent\textbf{Why Graph-based representations?}
From Table ~\ref{tab:TELSUM Ablation studies}, we can observe that either removal of any of the graph components ($GAE_{sent}$ or $GAE_{doc}$ or both) resulted in a significant drop in ROUGE scores in all the cases. Graph-based models tries to capture more relevant information with the help of neighboring nodes.
%\textcolor{red}{change wording?}.
Also, graph autoencoders further help us to achieve this in an effective and computation-friendly manner. 

% tries to capture more relevant and significant information present in a document.  \textcolor{red}{highlight the importance of GAE here}
% As mentioned above, the usage of GAE for document representations has improved ROUGE scores.  Using GAE only for sentence representations, but not for document representation has lower ROUGE scores than the state-of-art results. 
%\textbf{Why Contrastive Loss?}
\noindent\textbf{Importance of contrastive loss}
Contrastive loss~\cite{hadsell2006dimensionality} effectively operates on a pair of embeddings from similar vector space, specifying whether the two samples are “similar” or “dissimilar”. Hence, we minimize the contrastive loss between the graph-based document representation and the average of predicted top K sentences.

% Contrastive loss is calculated between the predicted summary and the document representation learned by GAE. Here, the predicted summary is the average of the top K sentences from the article. However, as we have mentioned in the ablation studies, the usage of GAE has improved ROUGE scores, as shown in Table~\ref{tab:TELSUM Ablation studies}. Using GAE only for sentence representations, but not for document representation has lower ROUGE scores than the state-of-art results.

% We observe that there is a significant drop in ROUGE scores, proving the importance of sentence position while generating summaries.

% \textcolor{red}{when we are saying seven indian languages, mention those seven}
\subsection{Performance on XL-sum (Seven Indian Languages)}
\vspace{-0.05cm}
We compare the performance of our model \textsc{GAE-ISumm} on the multilingual XL-sum~\cite{hasan2021xl} dataset extracted from BBC news.
%We observe that these summaries do not capture the complete information in the document. \textcolor{red}{above sentence continuity?}
To evaluate the model performance, we consider seven Indian language datasets  (``te, ta, gu, pa, bn, mr, hi'') from XL-sum and compare \textsc{GAE-ISumm} performance with the current state-of-art multilingual approach proposed by~\cite{dabre2021indicbart}.
IndicBERT  ~\cite{kakwani2020indicnlpsuite} is used as the feature extraction model to obtain the text representations, and we consider the top one (or) two scored sentences (according to the dataset average summary length) predicted by our model as summary.
Table~\ref{tab:XLSUM results only language} illustrates the results obtained on the XL-sum dataset. From Table~\ref{tab:XLSUM results only language}, we observe that \textsc{GAE-ISumm} outperformed the state-of-the-art models for four Indian languages (``gu, mr, te, and ta'').
In particular, our proposed model performed well on the Dravidian languages (``te, ta''). 
% and two of the Indo-Aryan languages (``gu, mr''). 
Although R-2 scores of \textsc{GAE-ISumm} showcase close performance to mBART and IndicBART for ``bn, hi, and pa'', however, \textsc{GAE-ISumm} report lower scores in the case of R-1 and R-L.
% , \textsc{GAE-ISUMM} reports lower performance (ROUGE-1 and ROUGE-L) for these three Indian languages.
% \textcolor{red}{modify sentence}
The lower performance can be mainly because the document size of XL-sum datasets ranges from 6 to 110 sentences. However, their gold summaries extracted are confined to one or two sentences (under-representation of the document). 
%Hence, the document summaries of XL-sum for many languages are under-representation of the document.
Overall, we compare our unsupervised model with their supervised multilingual setting and achieve competitive or better results.

% Figures~\ref{fig: XL-SUM example} shows examples of human-annotated summaries and the summary predicted by \textsc{GAE-ISumm} for Telugu.
% From Figures~\ref{fig: XL-SUM example}, we observe that in the generated summary,~\textsc{GAE-ISumm} extracts all the essential named entities, syntactic structure, and semantic information of the document.

% and the ROUGE-2 score of their multilingual model.

%However, the document size of XL-SUM ranges from 6-110 sentences, and the summaries extracted are confined to one or two sentences. 

\subsection{Does Cross-Lingual Models Improve Summarization Performance?}
Generally, multilingual pre-trained models are evaluated by their capacity for knowledge transfer across languages.
% This can be done either by training the summarization model on English data only or English+ other Indian language data using (for example) mBART representations. 
%\textcolor{red}{check the below sentence once}
To investigate the cross-lingual language transfer, we also experiment with our \textsc{GAE-ISumm} in bilingual settings on the XL-sum dataset. 
Bilingual summarization can be done either by training the model on a single language (high-resource language) alone and testing on the second language  (low-resource language) or by training on both languages and testing on the second language. This allows models to benefit from the high-resource languages. 
% To better assess the usefulness of the proposed dataset \textsc{TELSUM} and the existing multilingual dataset (XL-sum), we evaluate our \textsc{GAE-ISumm} in a cross-lingual setting.
We examine our \textsc{GAE-ISumm} in a cross-lingual environment to further evaluate the use of the proposed dataset \textsc{TELSUM} and the current multilingual dataset (XL-sum). 
% Here, we performed two cross-lingual settings:
% (i) we trained our model on Hindi and another Indian language (Telugu, Tamil, Bengali, Gujarati, Marathi) and tested on the corresponding Indian language, (ii) trained on English + another Indian language (Telugu, Tamil, Bengali, Gujarati, Marathi) and tested on the corresponding Indian language.
Here, we performed our cross-lingual experiments by considering ``Hindi (hi)'' and ``English (en)'' from XL-sum dataset as the high-resource languages and ``bn, gu, mr, pa, ta, and te'' to be the low-resource languages. We trained on both the high and low resource languages and tested on the low-resource languages. We also experimented by considering ``hi'' as the low-resource language and ``en'' as the high-resource language. From Table~\ref{tab:cross-lingual XLSUM results},  we observe that training in a bilingual setting has improved ROUGE scores.
The bilingual setting of ``hi'' with ``te, bn, gu, pa'' has improved ROUGE scores compared to the monolingual setting. Also, the bilingual setting of ``hi'' with other languages performed better than ``en'' due to the morphological and structural similarity of Indian languages.
We report the cross-lingual transfer results for~\textsc{TELSUM} dataset in Appendix (refer Table~\ref{tab:TELSUM Ablation studies}).
%\textcolor{red}{left task}
% \textcolor{red}{add something about similarities within Indian languages.}
%Table~\ref{tab:XLSUM results cross lingual english} reports the cross-lingual performance in two settings: (i) hi + remaining six Indian languages, and (ii) en + remaining seven Indian langauges. 
%Can we add something related to similarities present between the languages ? \textcolor{red}{change the wording}
%\subsubsection{Cross lingual summary experiments with English}
%Similar to above setting, here we trained our model on English and the remaining Indian languages( Hindi, Telugu, Tamil, Bengali, Guajrati, Punjabi, Marathi). \textcolor{red}{ablation studies for the table}
%\textcolor{red}{And mention that the number of sentences u considered as summary is based upon the dataset?}

\subsection{Performance on Other Datasets}

We employ our \textsc{GAE-ISumm} model on other existing summarization datasets of different Indian languages, as reported in Table~\ref{tab:Other Datasets results}.
%Table~\ref{tab:Other Datasets results} reports the performance of \textsc{GAE-ISumm} and compares it with the state-of-the-art model results. 
Observations from Table~\ref{tab:Other Datasets results} that \textsc{GAE-ISumm} yield superior performance on all the datasets with high ROUGE scores (R-1, R-2, and R-L).
Therefore, we argue that our \textsc{GAE-ISumm} model can be generalizable to any resource-poor language dataset.  %\textcolor{red}{change this}
%\textcolor{red}{Hindi Short has no existing values, should we mention that?}
% Moreover, we compare our unsupervised model with their supervised multilingual model and still achieve better results.
% a recently released dataset XL-SUM\cite{hasan-etal-2021-xl}. XL-SUM
% released summarization data on multiple languages where both thmoe document and summary are extracted from BBC news\footnote{https://www.bbc.co.uk/ws/languages}. 
%\textcolor{red}{anything to add about english :/?}

\section{Conclusion and Future Work}
This study addresses the robust representational capabilities of graph-based techniques, RNNs, and neural networks by proposing a unique unsupervised summarization model. Using cluster representations from GRU and GAE, we create a model that learns text representations and a document summary.
The positions, relevance, and semantic value of the sentences are also considered while ranking them in our approach.
Our experiments showcase that \textsc{GAE-ISumm} outperforms all baseline models and provide benchmark results on the \textsc{TELSUM}.
With the help of existing summarization datasets, we examine the effectiveness of our model for other Indian languages in both monolingual and bilingual settings.

In the future, we plan to introduce an abstractive model taking summarization a step forward in low-resource Indian languages. 
We also aim to expand our model to other non-Indian language datasets by adapting to diverse languages' syntactical and morphological features.
%\textcolor{red}{modify conclusion?}
% \vspace{-0.2cm}

\section{Ethical Statement}
\label{ethical_statement}
\vspace{-0.05cm}
We reused the publicly available datasets XL-sum  (\url{https://github.com/csebuetnlp/xl-sum}, BNLPC, NCTB~\cite{chowdhury2021unsupervised}, Marathi summarization dataset ~\url{https://github.com/pratikratadiya/marathi-news-document-dataset}, Hindi-short summarization dataset (~\url{https://www.kaggle.com/datasets/disisbig/hindi-text-short-summarization-corpus})  to compare our state-of-art models. We read the terms and conditions that these authors/publishers provided for the usage of these datasets and we see no harm.

\textbf{Fair Compensation}: 
We provided the data to \emph{Elancer IT Solutions Private Limited}~\footnote{http://elancerits.com/} company for getting the annotated summary.
In order to perform the annotation process, \emph{Elancer IT Solutions Private Limited} chose five native speakers of Telugu with excellent fluency. The company itself properly remunerates all the annotators.

\textbf{Privacy Concerns}: We have gone through the privacy policy of samyam website~\footnote{https://telugu.samayam.com/privacy-policy/privacypolicy/64302688.cms}. However, the copyrights of the articles remain with the publisher/authors ONLY. We do not foresee any harmful uses of using the data from the website.

\bibliography{ijcnn}
\bibliographystyle{IEEEtran}

\appendix 

\begin{table}[ht]
\vspace{0.2em}
\scriptsize
\centering
\caption{ Crosslingual experiments on \textsc{TELSUM} dataset. }
\label{tab:TELSUM Crosslingual experiments}
\resizebox{0.5\textwidth}{!}
{\begin{tabular}{|l|c c c|}
\hline
% \multicolumn{1}{|c|}{} & \multicolumn{3}{c|}{\textbf{Fine-Tuning}} \\ 
\cline{1-4} 

\textbf{Crosslingual language} & \textbf{R-1} & \textbf{R-2} & \textbf{R-L}      \\ \hline 
 
hi & 46.8& 37.9 & 44.2\\
en & 47.8 & 37.91 & 44.8\\
% Without ${score}_{pos}$ & 43.14 & 32.97 & 39.95 \\ 
\hline
%\multicolumn{4}{c}{ P = Precision, R = Recall, F1 = F1-score} \\
\end{tabular}}

\end{table} 

%\newpage

\begin{figure}[ht]
% \vspace{-0.5em}
%\scriptsize
\centering
% \textbf{Sentence} \\ \hline 
\includegraphics[width=\linewidth]{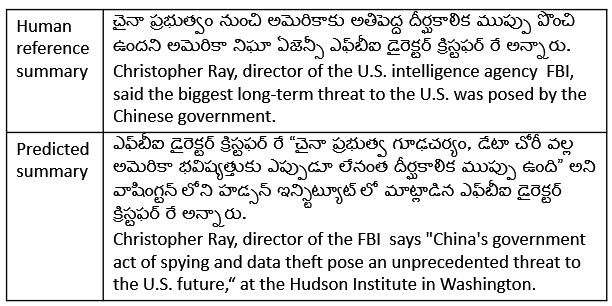}
\caption{Example of Actual and Predicted summary on XL-sum.}
\label{fig: XL-SUM example}
\end{figure}

\begin{figure}[t]
\vspace{-1em}
%\scriptsize
\centering
% \textbf{Sentence} \\ \hline 
\includegraphics[width=\linewidth]{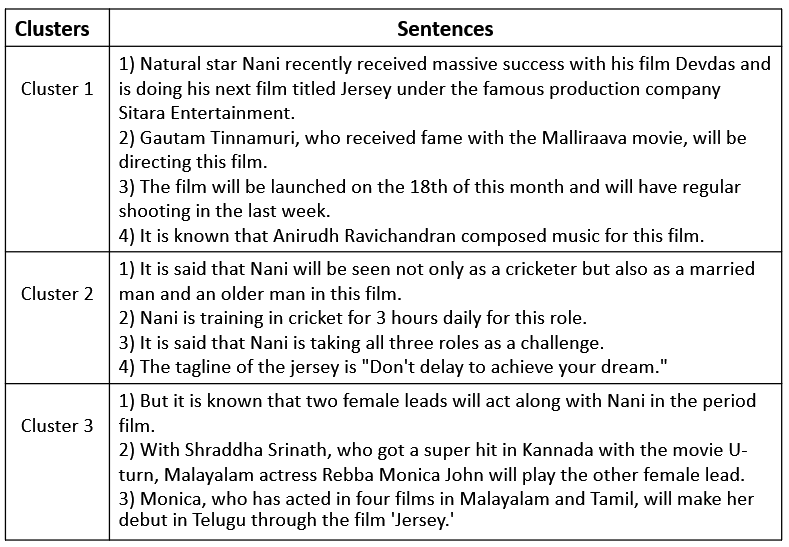}
\caption{Clusters formed by a sample article from the TELSUM dataset(English version).}
\label{fig: english cluster analysis}
\end{figure}

\end{document}